\documentclass[10pt,twocolumn,journal,letterpaper,review]{IEEEtran}
\usepackage{amsmath, amsthm, amssymb}
\usepackage{url,flushend,multirow,booktabs}
\usepackage[ruled,linesnumbered]{algorithm2e}  
\usepackage{graphicx}
\usepackage{bm}
\usepackage{cite}
\usepackage{subcaption}
\usepackage{float}
\usepackage{color}
\usepackage{multirow}
\usepackage{makecell}
\usepackage[colorlinks,citecolor=green,urlcolor=blue,bookmarks=false,hypertexnames=true]{hyperref}

\newcommand{\eg}{e.g.}
\newcommand{\ie}{i.e.}

\newcommand{\etal}{\textit{et al.}}

\DeclareFixedFont{\mf}{OT1}{ptm}{m}{n}{10pt}
\DeclareFixedFont{\mfb}{OT1}{ptm}{bx}{n}{10pt}

\begin{document}
%

\title{Sample-level Adaptive Knowledge Distillation for Action Recognition}
%
%
%

\author{Ping~Li, Chenhao Ping, Wenxiao Wang, and Mingli Song
\thanks{Manuscript received Nov 15th, 2024.}
}

\markboth{Draft}
{LI \MakeLowercase{\textit{et al.}}:~Sample-level Adaptive Knowledge Distillation for Action Recognition}
%

\maketitle

\begin{abstract}
Knowledge Distillation (KD) compresses neural networks by learning a small network (student) via transferring knowledge from a pre-trained large network (teacher). Many endeavours have been devoted to the image domain, while few works focus on video analysis which desires training much larger model making it be hardly deployed in resource-limited devices. However, traditional methods neglect two important problems, i.e., 1) Since the capacity gap between the teacher and the student exists, some knowledge w.r.t. difficult-to-transfer samples cannot be correctly transferred, or even badly affects the final performance of student, and 2) As training progresses, difficult-to-transfer samples may become easier to learn, and vice versa. To alleviate the two problems, we propose a \textbf{S}ample-level \textbf{A}daptive \textbf{K}nowledge \textbf{D}istillation (\textbf{SAKD}) framework for action recognition. In particular, it mainly consists of the sample distillation difficulty evaluation module and the sample adaptive distillation module. The former applies the temporal interruption to frames, i.e., randomly dropout or shuffle the frames during training, which increases the learning difficulty of samples during distillation, so as to better discriminate their distillation difficulty. The latter module adaptively adjusts distillation ratio at sample level, such that KD loss dominates the training with easy-to-transfer samples while vanilla loss dominates that with difficult-to-transfer samples. More importantly, we only select those samples with both low distillation difficulty and high diversity to train the student model for reducing computational cost. Experimental results on two video benchmarks and one image benchmark demonstrate the superiority of the proposed method by striking a good balance between performance and efficiency. 
\end{abstract}

\begin{IEEEkeywords}
Knowledge distillation; action recognition; model compression; sample-level distillation.
\end{IEEEkeywords}

 \ifCLASSOPTIONpeerreview
 \begin{center} \bfseries EDICS Category: 3-BBND \end{center}
 \fi

\IEEEpeerreviewmaketitle

\section{Introduction}
\label{sec:intro}  
\IEEEPARstart{D}{eep} neural networks \cite{krizhevsky-nips2012-imagenet, vaswani-nips2017-attention} have exhibited its great power in image and video analysis. However, their model size grows up rapidly as the advanced technologies are updated continuously, and this problem becomes more serious in video analysis including action recognition as a fundamental task, leading to heavy computational cost and much difficulty in deploying them on resource-constrained devices. Hence, compressing neural networks is very necessary in highly-demanding environment, and Knowledge Distillation (KD) \cite{geoffrey-arxiv2015-kd} has established itself as a useful model compression technique. Essentially, it transfers knowledge (\eg, outputs, gradients, intermediate features) from a powerful large network (\ie, teacher) to a lightweight small network (\ie, student) learned on training samples.

\begin{figure}[!t]
	\centering
	\includegraphics[width=0.88\linewidth]{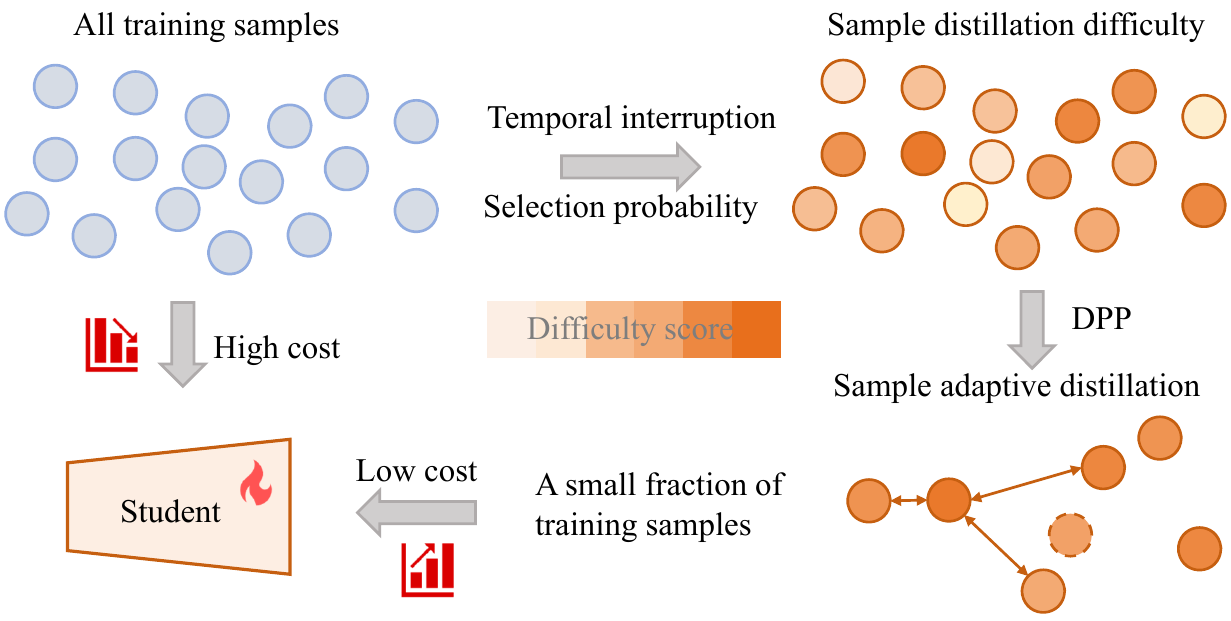}
	\caption{Illustration of motivation. We first determine the difficulties of all samples by adopting the temporal interruption strategy and considering the sample selection probability. Then, only a small fraction of training samples are chosen in each epoch for a much lower cost and a better student performance.}
	\label{fig:motivation}
	\vspace{-4mm}
\end{figure}

While KD methods \cite{geoffrey-arxiv2015-kd, pham-wacv2024-frequency, zong-iclr2023-kd} have been widely used in image analysis, there are only very few works exploring KD for action recognition. Some early works \cite{crasto-cvpr2019-mars, stroud-wacv2020-d3d} adopt cross-modal distillation by transferring knowledge from optical-flow network (teacher) to RGB network (student), and one recent work \cite{wang-aaai2024-gkd} adopts a pre-trained generative model (\ie, conditional variational autoencoder) for distilling attention-based feature knowledge. These methods all explore the early Convolutional Neural Network (CNN)-based action recognition models \cite{ji-tpami2012-3d, simon-nips2014-twostream} using both RGB and optical flow features, which is time-consuming, and they fail to consider the temporal dynamics in distillation. Importantly, they overlook \textbf{two} critical facts, \ie, \textbf{1)} Since the capacity gap between the teacher and the student exists, some knowledge \textit{w.r.t.} difficult-to-transfer samples cannot be correctly transferred, or even badly affects the final performance of student; and \textbf{2)} As training progresses, difficult-to-transfer samples may become easier to learn, and vice versa. Meanwhile, when all samples are used during student training, it incurs large computational overheads. So it is necessary to select those samples more benefiting distillation to save cost, as illustrated in Fig.~\ref{fig:motivation}. For example, we choose only 10\% samples to achieve comparable or even better performance than previous KD methods using all samples with SlowFast \cite{feichtenhofer-iccv2019-slowfast} model on UCF101 \cite{soomro-arxiv2012-ucf101} dataset, while the training time is greatly reduced to only one-fourth of previous ones.

This work addresses the above issues from sample-level distillation perspective, and introduces two concepts, \ie, \textit{sample distillation difficulty} and \textit{sample distillation strength}. The former stems from a fact that different samples have varying difficulty in transferring the knowledge from teacher. The latter determines to what degree the KD loss should be more emphasized at sample level during student learning, and it has negative correlation with sample distillation difficulty.

Regarding the first problem, we develop a sample distillation difficulty evaluation module which involves a temporal-dynamic interruption strategy. It adopts the dropout-shuffle technique that dynamically adjusts the interruption ratio of frames. Note that it can easily adapt to image analysis by conducting dropout-shuffle along the spatial dimension at pixel level. Essentially, it increases the learning difficulty (\eg, missing key frames, temporal misalignment among frames from different actions) of samples to enlarge the distillation loss between teacher and student. Usually, the larger the loss value, the lower the distillation difficulty is. This is because there exists large capacity difference to be diminished by distillation. Meanwhile, those samples selected multiples times are expected to have a low probability to be selected in future, which means these samples are difficult to transfer knowledge and should be less emphasized during distillation. The sample selection probability and the distillation loss value are coupled to determine the sample distillation difficulty.

Regarding the second problem, we design a sample adaptive distillation module, which dynamically calculates the sample distillation strength, according to the interruption ratio and the distillation difficulty. In particular, those highly difficult-to-transfer samples correspond to lower distillation strength and should learn more from ground-truth by vanilla loss. Meanwhile, difficult-to-transfer samples in early training may become easier in later training, since student learns more knowledge from teacher as training progresses. Hence, difficult-to-transfer samples may become less while easy-to-transfer ones become more during distillation. Moreover, we adopt the Determinantal Point Process (DPP) \cite{chen-nips2018-greedy} sampling to select those diverse samples with low distillation difficulty to make the student more robust. 

Overall, we propose a \textit{Sample-level Adaptive Knowledge Distillation} (\textbf{SAKD}) framework for action recognition. It is an easy-to-use plug-in technique and can be easily embedded into existing KD methods. Extensive experiments on two video benchmarks, \ie, UCF101 \cite{soomro-arxiv2012-ucf101} and Kinetics-400 \cite{kay-arXiv2017-kinetics}, and one image benchmark CIFAR-100 \cite{wah-2011-cifar}, have validated both effectiveness and efficiency of our approach. 

\section{Related Work}
\label{sec:relatedwork}
This section briefly reviews the related works in knowledge distillation and action recognition.

\subsection{Knowledge Distillation}
Knowledge Distillation (KD) reduces the inference model size without sacrificing the performance by transferring knowledge from teacher to student. Previous methods can be roughly divided into two groups, \ie, logit-based \cite{geoffrey-arxiv2015-kd}\cite{li-aaai2023-ctkd}\cite{sun-cvpr2024-lskd} ones and feature-based \cite{han-nips2015-learning}\cite{liu-iclr2023-norm}\cite{pham-wacv2024-frequency} ones. This allows student model to be deployed on resource-constrained devices efficiently. 

\textbf{Logit-based KD}. Student imitates the output logits of teacher. Hinton \etal~\cite{geoffrey-arxiv2015-kd} developed the first KD method by transferring the output distributions via soft labels from teacher to student. There are several attempts to better handle the logits, \eg, Yang \etal~\cite{yang-iccv2023-unifiedkd} normalizes the non-target logits from both teacher and student to use the soft labels for distillation; Wu~\etal~\cite{wu-eccv2022-tinyvit} sparsify the logits of teacher and store them in disk in advance to save the memory and computation cost, resulting in a tiny ViT (Vision Transformer) as student. To narrow down the gap of outputs, Li \etal~\cite{li-aaai2023-ctkd} employ a dynamic and learnable temperature to control the task difficulty level during student learning but the model is difficult to converge; Sun \etal~\cite{sun-cvpr2024-lskd} set the temperature as the weighted standard deviation of logit and perform a plug-and-play Z-score preprocess of logit standardization, which facilitates only logit-based KD methods.

\textbf{Feature-based KD}. It makes the intermediate feature of student approach that of teacher \cite{liu-iclr2023-norm}. For example, Han \etal \cite{han-nips2015-learning} employ one layer feature of teacher as the target of student and compute the Euclidean distance of their features. Afterwards, some works attempt to transfer knowledge by activation map \cite{zagoruyko-iclr2017-attentionkd}, feature distribution \cite{passalis-eccv2018-learning}, and pairwise similarity \cite{tung-iccv2019-similarity}, but these early works fail to consider the dynamics of the relation between teacher and student. To simplify the knowledge of teacher model, Wang \etal~\cite{wang-tip2024-dkd} propose a dual KD framework to reduce the side effects of teacher and utilize the optimal transport distance to measure the difference of feature maps between teacher and student. In addition, Zhu \etal~\cite{zhu-nips2022-kd}  verify the existence of undistillable classes by illustrating their correlation with capacity gap between teacher and student, the similarity of which is used to reflect the distillability level of student model.

\subsection{Action Recognition}
Action recognition model assigns class label to the video involving one action, \eg, \textit{running}. Early works adopt CNNs as the backbone, either using two-stream network \cite{simon-nips2014-twostream, wang-eccv2016-action} to extract both RGB features along the spatial dimension and optical flow features along the temporal dimension, or using 3D CNNs \cite{carreira-cvpr2017-kinetics, ji-tpami2012-3d, tran-cvpr2015-learning, yang-cvpr2020-temporal}  to capture the spatiotemporal features. Taking advantage of two-stream network and 3D CNNs, Feichtenhofer \etal~\cite{feichtenhofer-iccv2019-slowfast} combine them together and design the typical SlowFast network, where a slow pathway captures spatial semantics and a fast pathway captures motion at fine temporal resolution. Recent works \cite{dosovitskiy-iclr2021-transformers, li-cvpr2023-unmasked, liu-cvpr2022-video, wu-iccv2023-temporal}  apply Transformer \cite{vaswani-nips2017-attention} to the action recognition task by using the model pre-trained on large-scale dataset. More recently, Wang \etal~\cite{wang-aaai2024-gkd} present the generative feature KD framework to train student, which employs the conditional variational auto-encoder to extract attention-based features, but it brings about additional cost incurred by auto-encoder.

\section{Method}
\label{sec:method}
This section introduces the framework of Sample-level Adaptive KD in action recognition task, and describes the principles of the sample distillation difficulty evaluation and adaptive distillation strategy.
\begin{figure*}[!t]
	\centering
	\includegraphics[width=0.88\linewidth]{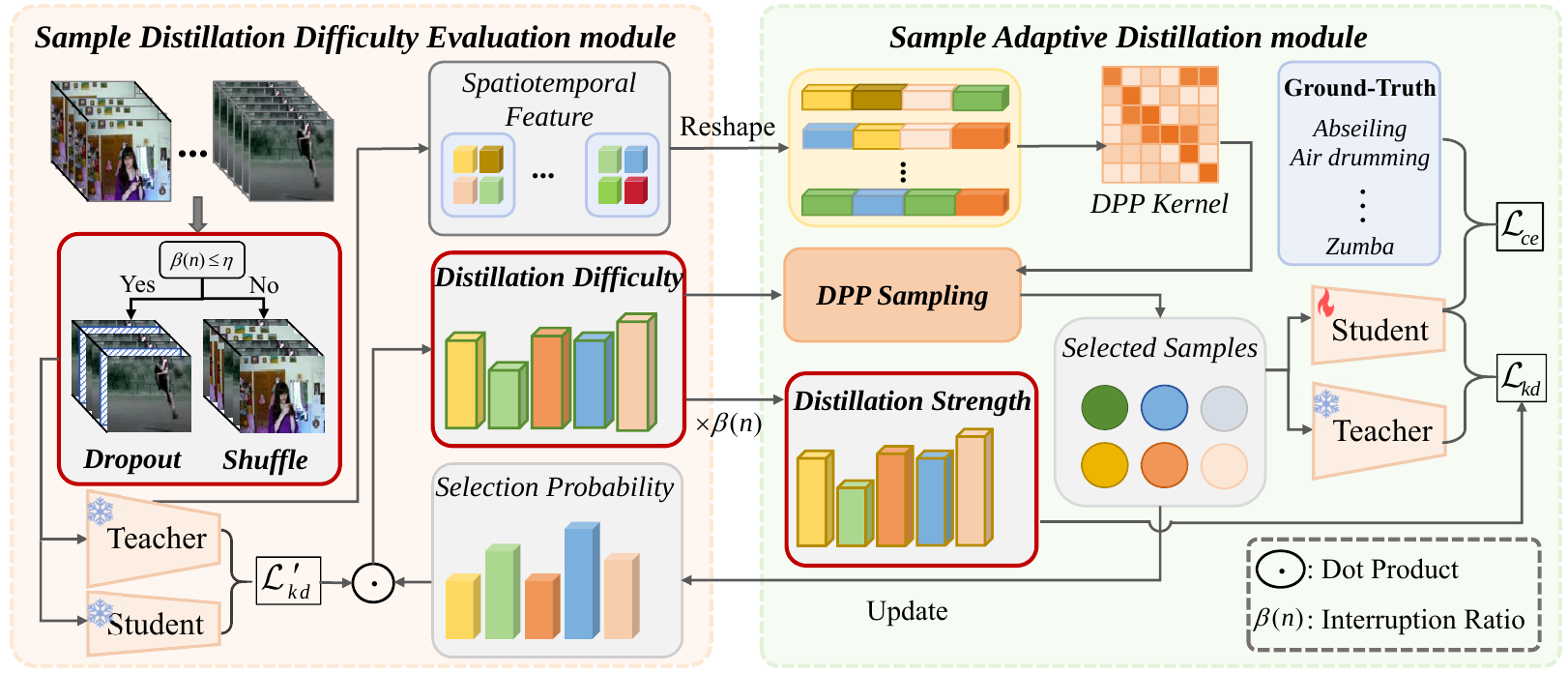}
	\caption{Overall framework of the Sample-level Adaptive Distillation (SAKD) approach.}
	\label{fig:framework}
	\vspace{-3mm}
\end{figure*}

\subsection{Problem Definition and Preliminary}  
KD learns a small network (\ie, student $\mathcal{S}$) by transferring the knowledge from a large network (\ie, teacher $\mathcal{T}$), where the latter is usually a pre-trained model with rich knowledge. Action recognition models are mostly very large and hardly to be deployed in resource-limited devices. Given a source video dataset, we divide it into training set $\mathcal{D}_{train} = \{\mathcal{V}_i, y_i\}_{i=1}^N$ and test set $\mathcal{D}_{test}= \{\mathcal{V}'_{i'}, y'_{i'}\}_{i'=1}^{N'}$, where $\mathcal{V}$ is video sample and $y\in \mathbb{R}$ is its action label from $K$ categories. The random sampling is operated on each video to obtain $T$ frames. For the $i$-th video clip, its RGB frames with height $H$ and width $W$ are stacked in a tensor $\mathbf{X}_i\in \mathbb{R}^{3\times T\times H\times W}$. Thus, it yields a training set $\mathcal{X} = \{\mathbf{X}_i, y_i\}_{i=1}^N$ for student learning and a test set $\mathcal{X}' = \{\mathbf{X}'_{i'}, y'_{i'}\}_{{i'}=1}^{N'}$ for inference.

During training, the video clips in $\mathcal{X}$ are input to both teacher $\mathcal{T}$ and student $\mathcal{S}$, which adopt the commonly used backbone (\eg, ResNet \cite{he-cvpr2016-residual}) for action recognition models such as Slowfast \cite{feichtenhofer-iccv2019-slowfast} and Temporal Pyramid Network \cite{yang-cvpr2020-temporal} to derive the spatiotemporal feature maps, \ie, $\mathbf{F}\in \mathbb{R}^{N\times c\times h\times w}$, where $\{c, h, w\}$ denote channel, height, and width. The dimensions of feature maps $\mathbf{F}^{\mathcal{T}}$, $\mathbf{F}^{\mathcal{S}}$ of both teacher and student are forced to be the same by a mapping function $f(\cdot)$, \eg, 1$\times$1 convolution or auto-encoder. 

For feature-based KD, the distillation loss minimizes the gap of features between teacher and student to make student imitate teacher, \ie,
\begin{equation}
	\label{eq:kdloss_fea}
	\mathcal{L}_{kd}^{fea} =  \frac{1}{N}\sum_{i=1}^N \|f(\mathbf{F}_i^{\mathcal{T}}) - f(\mathbf{F}_i^{\mathcal{S}}) \|_2^2,
\end{equation}
where $\|\cdot\|_2$ denotes the $\ell_2$-norm. 

For logit-based KD, the distillation loss minimizes the discrepancy between the output probabilities of teacher and student, \ie,
\begin{equation}
	\label{eq:kdloss_logit}
	\mathcal{L}_{kd}^{logit}  = \frac{1}{N}\sum_{i=1}^N \tau^2 KL(\sigma(q_i^{\mathcal{T}}/\tau), \sigma(q_i^{\mathcal{S}}/\tau)),
\end{equation}
where $KL(\cdot)$ is the Kullback-Leibler divergence, $\{q^{\mathcal{T}}, q^{\mathcal{S}}\}$ are the logits of teacher and student respectively, $\sigma(\cdot)$ is the softmax function, and $\tau$ (set to 4 \cite{geoffrey-arxiv2015-kd}) is the temperature to scale the smoothness of two probability distributions.

The vanilla loss of action recognition model adopts the Cross-Entropy (CE) loss, \ie, $\mathcal{L}_{ce} = -\frac{1}{N}\sum_{i=1}^N y_i \log q_i^{\mathcal{S}}$.

\subsection{Overall Framework}
The overall framework of Sample-level Adaptive Knowledge Distillation (SAKD) for action recognition is illustrated in Fig.~\ref{fig:framework}, which is composed of the Sample Distillation Difficulty (SDD) evaluation module and the Sample Adaptive Distillation (SAD) module. Unlike previous works using all to learn student model, SAKD selects only a small fraction of training samples for optimizing the model to save the cost. Essentially, we first evaluate the distillation difficulty of all training samples, and then select a small fraction of easy-to-transfer samples with diversity to train student model. Those selected samples form a selection subset $\mathcal{D}_{sel}$ dynamically in every epoch. The working mechanism of SAKD is described below. For example, we empirically found that the performance (\eg, accuracy) of our method using only 10\% samples is comparable or even surpasses that of previous KD methods using all samples, and the training time is greatly reduced to only one-fourth of the previous ones. 

First, the video samples in $\mathcal{D}_{train}$ are fed into the sample distillation difficulty evaluation module to compute the distillation difficulty score $\zeta$ for each sample. Note that the dropout-shuffle strategy is applied to video frames as a temporal interruption skill to enhance samples for increasing their learning difficulty, such that the distillation loss is enlarged for better evaluating distillation difficulty. These samples are fed into both teacher and student to derive the spatiotemporal features $\{\mathbf{F}^{\mathcal{T}}, \mathbf{F}^{\mathcal{S}}\}$ or output logits $\{q^{\mathcal{T}}, q^{\mathcal{S}}\}$ by forward propagation. They are used to calculate the distillation loss $\mathcal{L}_{kd}^{fea}$ or $\mathcal{L}_{kd}^{logit}$ using all samples. The loss value and the sample selection rate (\ie, historical selection times of samples) are used to evaluate the distillation difficulty at sample level. Moreover, we adopt the DPP \cite{chen-nips2018-greedy} to evaluate the diversity among the selected samples, and choose those diverse samples with low distillation difficulty to participate in training student model. In addition, we estimate the sample distillation strength according to the distillation difficulty and the interruption ratio per epoch, for better transferring knowledge from teacher to student by regarding it as an adaptive weighting ratio of the distillation loss and the vanilla loss.

\subsection{Sample Distillation Difficulty Evaluation}
Previous works treat all samples equally during distillation, while neglecting a fact that it may be trapped into an adverse situation, where both difficult-to-transfer and easy-to-transfer samples are emphasized equally which prevents more knowledge being transferred from teacher. We call this phenomenon ``\textit{distillation bottleneck}''. The reason behind this is different samples contribute differently during distillation due to the large gap between teacher and student. It is more difficult for some samples to perform knowledge transfer from teacher while it is easier for other samples. If all samples are equally treated, the student performance may saturate after certain training epoch, since the majority of samples are likely to be regarded as easy-to-transfer ones by student during later training. This inspires us to evaluate the sample-level distillation difficulty by less emphasizing the difficult-to-transfer samples and more emphasizing the easy-to-transfer samples per epoch. Here, we discriminate difficult-to-transfer samples in two aspects: 1) if the sample selection probability in the next epoch is low, student may hardly learn more knowledge from teacher with more epochs; 2) if the distillation loss difference between neighbouring epochs is small but the loss value is large due to the teacher-student gap, or the distillation loss itself in current epoch is small, this suggests the knowledge is difficult to transfer for this sample.

Mathematically, we define the sample distillation difficulty $\zeta_i$ as:
\begin{equation}
	\label{eq:distill_difficulty}
	\zeta_i = 1/(\textbf{p}^{sel}_i\cdot \tilde{\mathcal{L}}_i^{kd}).
	\vspace{-1mm}
\end{equation} 
where $\mathbf{p}^{sel}_i$ denotes the sample selection probability, and $\tilde{\mathcal{L}}_i^{kd}$ is the distillation loss on the enhanced samples by the dynamic-temporal interruption strategy, which dynamically does the dropout-shuffle operations on video samples. This adds the learning difficulty and diversifies samples during distillation. Details are shown below.

Given the $i$-th video sample $\mathbf{X}_i\in \mathbb{R}^{3\times T\times H\times W}$, we randomly dropout some frames or shuffle frames at some interruption rate $\beta(\cdot)\in (0, 1]$, which is a function of the current epoch number $n$, \ie, 
\begin{equation}
	\label{eq:dropout_rate}
	\beta(n) = 1-(1-n/N_{epoch})^{\theta},
\end{equation}
where $N_{epoch}$ is the maximum epoch, and the power $\theta$ is set to 0.9 as indicated by poly learning rate \cite{wu-tist2023-poly}. Here we introduce a threshold $\eta\in (0, 1]$ to control whether adopting the dropout or shuffle operation. When the interruption rate is less than $\eta$, we randomly dropout a percentage $\beta(n)$ of the frames in a video along the temporal dimension; when it is larger than or equal to $\eta$, we randomly shuffle a percentage $\eta$ of the frames in a batch, where there exist different actions. Motivated by the curriculum learning \cite{li-aaai2023-ctkd}, the interruption rate function adopts the polynomial learning rate policy. In another word, the interruption on frames is weak at early training and is gradually enhanced to increase the training difficulty of samples. 

By the dynamic interruption operation, we derive the enhanced video sample $\tilde{\mathbf{X}_i}$, which is fed into teacher model and student model to compute the distillation loss $\tilde{\mathcal{L}}_i^{kd}$. This loss value acts as an important factor that influences the evaluation of sample distillation difficulty. Usually, the larger the loss value, the lower the distillation difficulty is. The sample with lower distillation difficulty has a high probability to be selected for learning the student model. Meanwhile, those samples ever selected multiple times in different epochs are expected to have a low probability to be selected again, such that more different samples have the opportunity of participating in the student learning. This will also improve the generalization ability of student model during inference. To achieve this, we define the sample selection probability $\mathbf{p}^{sel}_i$ as: 
\begin{equation}
	\label{eq:sel_prob}
	\mathbf{p}^{sel}_i = 1/(\mathbf{\omega}_i + \epsilon),
\end{equation}
where $\mathbf{\omega}_i$ is the selection times of the $i$-th sample during training, and the constant $\epsilon>0$ (set to 1 empirically) is a smoothing factor to avoid the zero value of denominator. The normalization is applied to the vector $\mathbf{\omega}$ to ensure the probability will not be too small due to a large epoch. This selection probability acts as a correction factor that influences the evaluation of sample distillation difficulty. 

At the early training stage, the difficult-to-transfer samples with too large loss values might be wrongly assigned with a low distillation difficulty value according to Eq.~(\ref{eq:distill_difficulty}). Hence, we weight the loss value by a correction factor, \ie, the sample selection probability $\mathbf{p}^{sel}_i$. It makes sense because the correction factor will become small when the sample is selected multiple times, thus weakening the impact of difficult-to-transfer samples with large loss. As the training goes on, the number of easy-to-transfer samples will increase since the learning ability of student model is boosted gradually with more epochs. In this situation, both teacher and student may yield similar logits or intermediate features, leading to small distillation loss value. So, the samples with too similar outputs from teacher and student should not be selected in future, because it is hard for student to learn richer knowledge from teacher with these samples any more. These samples are also regarded as difficult-to-transfer ones.

\subsection{Sample Adaptive Distillation Module}
Traditional methods always fix the ratio of the vanilla loss and the KD loss. Worse still, this ratio is identified by grid searching from some range of hyper-parameters, which is inefficient at expensive computational cost. To overcome this drawback, we propose the sample adaptive distillation strategy, which dynamically estimates the distillation strength of each sample according to the sample distillation difficulty, \ie, the above ratio is adaptively adjusted to the current student model. 

Formally, the distillation strength $\alpha_{i,n}$ of the $i$-the sample in the $n$-th epoch is defined as:
\begin{equation}
	\label{eq:distill_strength}
	\alpha_{i,n}=\lambda \alpha_{i,n-1} + (1-\lambda)\beta(n)/\zeta_i,
\end{equation}
where the first term is the distillation strength of the previous epoch, the second term directly affects the distillation strength, and the hyper-parameter $\lambda\in (0, 1)$ that balances the contribution of the two terms. The first term consider the previous distillation strength. The second term reveals two facts: 1) the distillation strength $\alpha$ gradually rises up as the sample interruption rate $\beta(n)$ in Eq.~(\ref{eq:dropout_rate}) grows monotonically with increasing epochs, and 2) the distillation strength $\alpha$ is inversely proportional to the distillation difficulty $\zeta$, which makes sense since difficult-to-transfer samples should be less emphasized during distillation.

Another problem to be solved is to adaptively select those easy-to-transfer samples and diverse samples per epoch. For the former, we resort to the sample distillation difficulty, \ie, the samples with low distillation difficulty are treated as easy-to-transfer samples. For the latter, we adopt the Determinantal Point Process (DPP) \cite{kulesza-ftml2012-dpp} to evaluate the diversity of samples. Specifically, we define a DPP kernel matrix $\mathbf{\Lambda}\in \mathbb{R}^{N\times N}$ that measures the similarity of feature matrices. When the column vectors are more dissimilar, their angles are greater and thus the determinant of the kernel matrix, $\mathop{\ln} \det(\mathbf{\Lambda})$, is larger. 

To obtain the above feature matrices, it goes through the following steps. First, we feed video sample $\tilde{\mathbf{X}}_i$ into teacher model to derive the feature map $\mathbf{F}_i\in\mathbb{R}^{c\times h\times w}$. Second, we fuse the feature maps of two neighbouring epochs to yield the fusion feature map $\tilde{\mathbf{F}}_i\in\mathbb{R}^{c\times h\times w}$, \ie,
\begin{equation}
	\label{eq:fuse_fea}
	\tilde{\mathbf{F}}_{i,n}=\lambda \tilde{\mathbf{F}}_{i,n-1} + (1-\lambda ) \mathbf{F}_i,
\end{equation}
where $\lambda\in (0, 1)$ is the same as in Eq.~(\ref{eq:distill_strength}) to balance the fusion ratio of the current epoch and the previous epoch. Here, we update the feature by considering that of previous one to avoid large fluctuations of feature during training, which helps to stabilize the subsequent DPP sampling process. The fusion feature map $\tilde{\mathbf{F}}_i$ is reshaped to $\hat{\mathbf{F}}_i \in \mathbb{R}^d $ by flattening the feature map to a vector at sample level, where $d=c\cdot h\cdot w$. Now we can compute the above DPP kernel matrix, \ie, $\mathbf{\Lambda} = \hat{\mathbf{F}}\cdot\hat{\mathbf{F}}^\top$, where $\hat{\mathbf{F}}\in \mathbb{R}^{N\times d}$ is the feature matrix of all samples. 

To achieve the goal of selecting those samples with both low distillation difficulty and high diversity, we optimize the following objective function, \ie,
\begin{equation}
	\label{eq:sample_select}
	\mathop{\max}_{\mathcal{A}} \gamma \sum_{j=1}^{N_{sel}} \frac{1}{\zeta_{proj(j)}} + (1-\gamma) \ln 
	\frac{\det(\mathbf{\Lambda}_{\mathcal{A}})}{\det(\mathbf{\Lambda} + \mathbf{I})},
\end{equation}
where $\mathbf{I}\in \mathbb{R}^{N\times N}$ is an identity matrix, $\mathcal{A}$ is an index set of those selected samples, $N_{sel} = r\cdot N$ is the sample selection number, $r$ is a selection ratio, and hyper-parameter $\gamma\in [0, 1]$ is used to tradeoff the former distillation difficulty term and the latter diversity term. Here, there exists one-to-one projection $proj(\cdot)$ between the new index $j\in \{1, 2, \ldots, N_{sel}\}$ in the selected sample set $\mathcal{A}$ and the index $i$ of the source sample, and vice versa, \ie, $i = proj(j)$. The matrix $\mathbf{\Lambda}_{\mathcal{A}}$ denotes the kernel matrix derived from the features of those selected samples, which forms the selection subset $\mathcal{D}_{sel}\subset \mathcal{D}_{train}$. We adopt the greedy algorithm \cite{chen-nips2018-greedy} to optimize the above objective function by Cholesky decomposition to reduce the computational complexity. 

\subsection{Loss Function}
For action recognition task, the total loss consists of the vanilla classification loss $\mathcal{L}_{ce}$ and the KD loss $\mathcal{L}_{kd}$, and it is calculated on those selected video clips per epoch, \ie,
\begin{equation}
	\mathcal{L}_{total} = \frac{1}{N_{sel}} \sum_{j=1}^{N_{sel}} (1-\alpha_{proj(j)})\mathcal{L}_j^{ce} + \alpha_{proj(j)} \mathcal{L}_j^{kd}.
\end{equation}
where $\alpha_{proj(j)} \in (0, 1)$ is a dynamic distillation strength of the $j$-th selected video clip per epoch, and $\mathcal{L}_j^{kd}$ takes the form of Eq. (\ref{eq:kdloss_fea}) or (\ref{eq:kdloss_logit}). During the early training, difficult-to-transfer samples are expected to be less emphasized during distillation and student should learn knowledge more from ground-truth labels than from teacher, \eg, sample distillation strength $\alpha$ is less than 0.5. As the training progresses, the learning ability of student model is boosted gradually. Therefore, those originally difficult-to-transfer samples may become easier to transfer during the later training, and simultaneously the sample distillation strength gets larger by dominating the distillation process.

\section{Experiment}
\label{sec:experiment}
All experiments were performed on a server equipped with four 11G GeForce 2080Ti graphics cards. The codes are compiled with PyTorch 1.7, Python 3.8, and CUDA 10.1.

\begin{table}[!t]
\centering
\caption{Statistics of data. ``K''/``N'' is class/sample number.}
\vspace{-2mm}
\label{tbl:dataset}
		\setlength{\tabcolsep}{0.6mm}{
			\begin{tabular}{lrr c rr c rr}
				\toprule[0.75pt]
				\multirow{2}{*}{Dataset} & \multicolumn{2}{c}{Training} & &\multicolumn{2}{c}{Validation } & & \multicolumn{2}{c}{Test} \\ 
				\cmidrule[0.5pt]{2-3}  \cmidrule[0.5pt]{5-6} \cmidrule[0.5pt]{8-9}
				& $K$	   & $N$ 		    && $K$	   & $N$   && $K$	   & $N$ \\
				\midrule[0.5pt]
				UCF101\cite{soomro-arxiv2012-ucf101}	   & 101	&9,537   	& &-&-           && 101	&3,783  \\
				Kinetics-400\cite{kay-arXiv2017-kinetics}  & 400    &234,619	& & 400 & 19,761 && -	&-      \\
				CIFAR-100\cite{wah-2011-cifar}   		   & 100	&50,000     & & -	&-       && 100	&10,000  \\
				\toprule[0.75pt]
			\end{tabular} 
	}
\end{table}

\subsection{Datasets}  
We conduct experiments on two video benchmarks including UCF101 \cite{soomro-arxiv2012-ucf101}\footnote{https://www.crcv.ucf.edu/data/UCF101.php} and Kinetics-400 \cite{kay-arXiv2017-kinetics}\footnote{https://deepmind.com/research/open-source/kinetics}, and one image benchmark CIFAR-100 \cite{wah-2011-cifar}\footnote{http://www.cs.toronto.edu/~kriz/cifar.html}. Statistics are in Table~\ref{tbl:dataset}. 

\textbf{UCF101}\footnote{https://www.crcv.ucf.edu/data/UCF101.php} \cite{soomro-arxiv2012-ucf101} consists of daily-life action videos collected from YouTube, covering 101 different categories with a total of 13,320 video clips, whose total duration is approximately 27 hours. The videos are organized into 25 groups by category, with each group containing 4 to 7 videos. The video resolution is 320$\times$240, and the frame rate is 25 fps. We use the official splits with all images uniformly cropped to a size of 224$\times$224.

\textbf{Kinetics-400}\footnote{https://deepmind.com/research/open-source/kinetics} \cite{kay-arXiv2017-kinetics} was originally released by DeepMind and contains video clips collected from YouTube, covering 400 different action categories. Each category has at least 400 videos, with each video clip lasting approximately 10 seconds. It includes 234,619 samples for the training set and 19,761 videos for the validation set, with all input images uniformly cropped to a size of 256 $\times$ 256.

\textbf{CIFAR-100}\footnote{http://www.cs.toronto.edu/~kriz/cifar.html} \cite{wah-2011-cifar} contains 100 categories, each of which has 600 RGB images of size 32$\times$32. The training set has 50,000 images and the test set has 10,000 images. All images are uniformly rescaled to 32$\times$32.

 Following previous works \cite{zong-iclr2023-kd,li-aaai2023-ctkd,xie-cvpr2023-dynamickd}, we adopt the commonly used Top-1 accuracy and Top-5 accuracy as the evaluation metrics. Also we report the elapsed time of each epoch to show the training efficiency. Top-1 or Top-5 accuracy evaluates those samples whose ground-truth class takes up the top or one of the five leading positions of the candidate class set. 

\begin{table}[!t]
	\centering
	\caption{Batch size and training epoch.}
	\vspace{-2mm}
	\label{tbl:batch_epoch_setting}
		\setlength{\tabcolsep}{0.4mm}{
				\begin{tabular}{lcc  ccc  cccc}
						\toprule[0.75pt]
						\multirow{4}{*}{Model} & \multicolumn{5}{c}{Batch Size}    & &\multirow{3.5}{2pt}{$N_{epoch}$ } & &\\ 
						\cmidrule[0.5pt]{2-6} 
						& \multicolumn{2}{c}{Our method} & &\multicolumn{2}{c}{Student training}   && &&\\
						\cmidrule[0.5pt]{2-3}\cmidrule[0.5pt]{5-6} \cmidrule[0.5pt]{8-9} 
						& UCF101	   &Kinetics  	& &	UCF101	   &Kinetics &&	UCF101	   &Kinetics &\\
						\midrule[0.5pt]
						SlowFast \cite{feichtenhofer-iccv2019-slowfast}  &96  &128&&32 &48 &&50   &200 &\\
						TPN \cite{yang-cvpr2020-temporal}                &24  &-  &&4  &-  &&50   &-&\\
						Video ST \cite{liu-cvpr2022-video}               &24  &48 &&4  &8  &&100  &200 &\\
						\toprule[0.75pt]
					\end{tabular} 
			}
\end{table}

\subsection{Experimental Setup}
\label{sec:exp_set}
For UCF101 \cite{soomro-arxiv2012-ucf101}, the action recognition model parameters are initialized with the model pre-trained on Kinetics-400 \cite{kay-arXiv2017-kinetics}. For UCF101 and Kinetics-400, the number of frames $T$  in a clip after sampling is 32 and 16, respectively. The Stochastic Gradient Descent optimizer is used with a momentum of 0.9. The initial learning rate for UCF101 and Kinetics-400 is set to 1e-2, with a weight decay factor of 1e-4 after each epoch. For Kinetics-400, following \cite{liu-cvpr2022-video}, both the tiny version (SwinT) and the small version (SwinS) of Video Swin Transformer (Video ST) \cite{liu-cvpr2022-video} are pre-trained on ImageNet, while models are randomly initialized. For CIFAR-100, following \cite{pham-wacv2024-frequency}, it has an initial learning rate of 0.1 and a decay factor of 0.1 applied at epochs 150, 180, and 210. During student training, the parameters of teacher are frozen and do not participate in back-propagation. For CIFAR-100 \cite{wah-2011-cifar}, the batch size of our method using frozen model is 1024, the batch size of student training is 256, and the training epoch is set to 240. The batch size and training epoch settings of two video datasets are summarized in Table~\ref{tbl:batch_epoch_setting}. 

To reduce the training cost, we select a small fraction of samples with low distillation difficulty and high diversity every epoch or every five epochs. For UCF101 and Kinetics-400, the sample selection ratio $r$ is 0.1, \ie, 10\% of all training samples are selected every time, and this rate is set to 0.5 for CIFAR-100 since images have much less redundancy than video. The hyper-parameters are set as follows: $\lambda$ of the distillation strength in Eq.~(\ref{eq:distill_strength}) and the feature fusion in Eq.~(\ref{eq:fuse_fea}) is set to 0.1, $\gamma$ of the sample selection using DPP in Eq.~(\ref{eq:sample_select}) is set to 0.5, and the interruption (dropout-shuffle) ratio threshold $\eta$ is set to 0.5.

\textbf{Inference}. Given a test video or image, we do the normalization before feeding them into the student model to output the estimated action class or image label.

\subsection{Compared Methods}
We compare our SAKD method with two groups of State-Of-The-Art (SOTA) KD methods: 1) \textit{logit-based} ones include KD (vanilla Knowledge Distillation) \cite{geoffrey-arxiv2015-kd}, DKD (Decoupled KD) \cite{zhao-cvpr2022-dkd}, and CTKD (Curriculum Temperature KD)\cite{li-aaai2023-ctkd}; 2) \textit{feature-based} ones include CrossKD (Cross-head KD)\cite{wang-cvpr2024-crosskd}, DualKD\cite{wang-tip2024-dkd}, and GKD (Generative model based KD) \cite{wang-aaai2024-gkd}. We use the source codes publicly available from the original papers, and we try the best to implement DualKD \cite{wang-tip2024-dkd} by ourself since its code is unavailable. Our code is available in the attached file.

To verify the generalization ability of our SAKD method on both video and image samples, we examine the compared KD methods on three typical action recognition models, including SlowFast \cite{feichtenhofer-iccv2019-slowfast}, TPN (Temporal Pyramid Network) \cite{yang-cvpr2020-temporal}, and VideoST (Video SwinTransformer) \cite{liu-cvpr2022-video}, as well as two typical image classification models (settings follows \cite{pham-wacv2024-frequency}), including ResNet (Residual Neural Network) \cite{he-cvpr2016-residual} and WRN (Wide ResNet) \cite{zagoruyko-bmvc2017-residual}. For SlowFast, we sample 16 or 8 or 4 frames in a clip when the step is set to 8 or 16, termed SF16x8 or SF8x8 or SF4x16. For TPN teacher or student, we sample 32 or 8 frames along the temporal dimension and scale up frames along the spatial dimension by a factor of 2 or 8, termed TPN-f32s2 or TPN-f8s8. For VideoST, SwinS is teacher whose total Transformer layer at Stage~3 is 18, while SwinT is student whose that layer number is 6. For SlowFast and TPN, teacher and student adopt ResNet101 and ResNet50 as the backbone respectively.

\begin{table}[!t]
	\centering
	\vspace{-3mm}
	\caption{Performance on UCF101 \cite{soomro-arxiv2012-ucf101} in terms of Top-1/5 Accuracy (\%) / training time (min). ``$r$'' is sample selection ratio.}
	\label{tbl:ucf101}
	\scalebox{0.85}{
		\setlength{\tabcolsep}{0.3mm}{
			\begin{tabular}{l l c c cc c c ccr c cc r}
				\toprule[0.75pt]
				\multirow{2}{*}{Method} & \multirow{2}{*}{Type} & \multirow{2}{*}{$r$} &&
				\multicolumn{3}{c}{SlowFast\cite{feichtenhofer-iccv2019-slowfast}}&& 
				\multicolumn{3}{c}{VideoST\cite{liu-cvpr2022-video}} && 
				\multicolumn{3}{c}{TPN\cite{yang-cvpr2020-temporal}}\\
				\cmidrule(lr){5-7} \cmidrule(lr){9-11} \cmidrule(lr){13-15}
				& & && Top1$\uparrow$  & Top5$\uparrow$  & min$\downarrow$ & & Top1$\uparrow$  & Top5$\uparrow$  & min$\downarrow$ && Top1$\uparrow$  & Top5$\uparrow$  & min$\downarrow$ \\
				\midrule[0.5pt]
				\multirow{2}{*}{-} & Teac. & 100\% && 93.61 & 98.64 & 10.2 && 92.18 & 98.01 & 22.2 & & 90.21 & 96.51 & 32.8\\
				& Stud. & 100\% && 85.15 & 95.80 & 6.2 && 86.17 & 97.03 & 18.1 && 82.23 & 93.24 & 20.7  \\
				\midrule[0.5pt]
				\multirow{3}{*}{\makecell[l]{KD\cite{geoffrey-arxiv2015-kd} \\ \small arXiv'15 }} 
				& Vani. & 100\% && 87.92 & 97.56 & 6.4&& \underline{89.38}  & 98.25 & 20.0  && 86.02 & 94.60 & 23.4  \\
				& Ours & 10\% && \bf90.73 & \bf99.20 & \underline{4.6}&& \bf{89.43} & \bf99.04 & \underline{11.1} && \bf87.15 & \underline{96.61} & \underline{11.3}  \\
				& Ours$^\ast$ & 10\% && \underline{89.88} & \underline{98.80} & \bf1.4& & 87.70 & \underline{98.72} & \bf3.9& & 85.55 & \bf97.01 & \bf4.9 \\
				\midrule[0.5pt]
				\multirow{3}{*}{\makecell[l]{DKD\cite{zhao-cvpr2022-dkd} \\ \small CVPR'22}} 
				& Vani. & 100\% && \underline{90.50} & 98.46 & 6.8 & & \underline{91.60} & 98.43 & 19.9& & 85.40 & 94.05 & 23.5\\
				& Ours & 10\% && \bf91.07 & \bf99.31 & \underline{4.6} && \bf91.78 & \bf99.06 & \underline{12.2} & & \bf87.92 & \bf97.59 & \underline{13.0}\\
				& Ours$^\ast$ & 10\% && 89.93 & \underline{98.83} & \bf1.5& & 91.44 & \underline{98.72} & \bf4.0&& \underline{86.20} & \underline{96.88} & \bf5.0  \\
				\midrule[0.5pt]
				\multirow{3}{*}{\makecell[l]{CTKD\cite{li-aaai2023-ctkd} \\ \small AAAI'23}} 
				& Vani. & 100\% && 87.12 & 96.45 & 6.9& & \underline{89.61} & \underline{97.93} & 20.1 & & \underline{86.12} & 94.57 & 24.2\\
				& Ours & 10\% && \bf90.67 & \underline{98.88} & \underline{4.8} && \bf89.81 & \bf99.08 & \underline{12.1} & & \bf86.65 & \underline{96.84} & \underline{11.5}\\
				& Ours$^\ast$ & 10\% & & \underline{89.61} & \bf98.91 & \bf1.4& & 87.83 & 97.82 & \bf4.0& & 85.96 & \bf97.10 & \bf4.9\\
				\midrule[0.5pt]
				\multirow{3}{*}{\makecell[l]{GKD\cite{wang-aaai2024-gkd} \\ \small AAAI'24}} 
				& Vani. & 100\% && 86.83 & 96.03 & 7.0 && 88.24 & 97.97 & 20.8& & 86.37 & 94.63 & 24.9 \\
				& Ours & 10\% && \bf89.92 & \bf98.51 & \underline{5.3} && \bf88.42 & \bf98.16 & \underline{13.8} & & \bf88.72 & \bf97.28 & \underline{13.9}\\
				& Ours$^\ast$ & 10\% && \underline{87.92} & \underline{98.34} & \bf1.8& & \underline{87.23} & \underline{98.03} & \bf4.6&& \underline{87.64} & \underline{97.04} & \bf4.8  \\
				\midrule[0.5pt]
				\multirow{3}{*}{\makecell[l]{\small CrossKD\cite{wang-cvpr2024-crosskd}\\ \small CVPR'24}} 
				& Vani. & 100\% && \underline{90.79} & \underline{99.10} & 6.5&& \underline{87.63} & \underline{98.06} & 19.8 & & \underline{88.05} & 97.51 & 23.2 \\
				& Ours & 10\% && \bf90.82 & 98.91 & \underline{4.9}&& \bf{87.98} & \bf98.38 & \underline{11.6} & & \bf88.55 & \bf97.96 & \underline{12.0} \\
				& Ours$^\ast$ & 10\% && 90.04 & \bf99.23 & \bf1.5  & & 86.94 & 97.16 & \bf3.8&& 87.69 & \underline{97.29} & \bf4.9\\
				\midrule[0.5pt]
				\multirow{3}{*}{\makecell[l]{\small DualKD\cite{wang-tip2024-dkd} \\  \small TIP'24}} 
				& Vani. & 100\% && 86.65 & 96.82 & 7.0& & \underline{89.27} & \underline{98.06} & 20.5 && \underline{86.17} & 94.75 & 26.2 \\
				& Ours & 10\% && \bf90.32 & \bf98.79 & \underline{5.0} && \bf{89.58} & \bf99.87 & \underline{13.0} && \bf86.67 & \bf97.21 & \underline{11.4} \\
				& Ours$^\ast$ & 10\% && \underline{89.64} & \underline{98.72} & \bf1.6 & & 87.95 & 97.72 & \bf4.3& & 85.86 & \underline{96.76} & \bf5.0\\
				\toprule[0.75pt]
			\end{tabular}
		}
	}
\end{table}

\subsection{Quantitative Results}
We apply all compared KD methods on action recognition models, whose results are reported in Table~\ref{tbl:ucf101} (UCF101 \cite{soomro-arxiv2012-ucf101}) and Table~\ref{tbl:kinetics} (Kinetics-400 \cite{kay-arXiv2017-kinetics}). The results on image models are reported in Table~\ref{tbl:cifar} (CIFAR-100 \cite{wah-2011-cifar}). The best records are highlighted in boldface, and the second best ones are underlined. Here, ``Vanilla'' (Vani.) denotes existing KD method, ``Ours'' or ``Ours$^\ast$'' denotes our method selecting samples every epoch or every five epochs. 

From Table~\ref{tbl:ucf101} and Table~\ref{tbl:kinetics}, we observe that our SAKD method as a plug-in technique consistently improves the performance of three typical action recognition models including SlowFast \cite{feichtenhofer-iccv2019-slowfast}, VideoST \cite{liu-cvpr2022-video}, and TPN \cite{yang-cvpr2020-temporal} across six SOTA KD methods on UCF101 and Kinetics-400. Compared to the baseline (row~1) without knowledge distillation, both vanilla KD methods and ours perform better, which supports the claim that student indeed learns knowledge from teacher by KD. Meanwhile, vanilla KD methods consumes more time than the baseline due to the additional overheads incurred by the distillation module. On the contrary, our method greatly reduces training time, \eg, from 4.8 min to 1.5 min with SlowFast on CTKD method at a faster training speed by over three times. Moreover, our SAKD method outperforms the vanilla KD ones by using only 10\% of training samples. This is because ours takes into account the different distillation difficulty of samples and adaptively adjusts the distillation strength for each sample as training progresses, which allows the student to learn more knowledge from teacher. Especially, those difficult-to-transfer samples receive more emphasis during student training, further improving the performance. 

In addition, similar performance improvements can be found on CIFAR-100 as shown in Table~\ref{tbl:cifar}, which reports the results of six different backbone pairs in two types including WRN \cite{zagoruyko-bmvc2017-residual} and ResNet \cite{he-cvpr2016-residual}. While KD methods bring about improvements on the baseline, the gain is smaller using 50\% of training samples compared with that on video. This might because that CIFAR-100 is an image database with few redundancy and it may lose some representative images when applying the dropout strategy. This is a bit different from that of video with much redundancy.

\begin{table}[!t]
	\centering
	\caption{Performance on Kinetics-400 \cite{kay-arXiv2017-kinetics} in terms of Top-1/5 Accuracy (\%) / training time (hour). ``$r$'' is sample selection ratio.} 
	\label{tbl:kinetics}
	\scalebox{0.9}{
		\setlength{\tabcolsep}{1.0mm}{
			\begin{tabular}{l l c c c r c c r}
				\toprule[0.75pt]
				\multirow{2}{*}{Method} & \multirow{2}{*}{Type} & \multirow{2}{*}{$r$} &
				\multicolumn{3}{c}{SlowFast\cite{feichtenhofer-iccv2019-slowfast}} & \multicolumn{3}{c}{VideoST\cite{liu-cvpr2022-video}} \\
				\cmidrule(lr){4-6} \cmidrule(lr){7-9}
				& & & Top1$\uparrow$  & Top5$\uparrow$  & hr$\downarrow$ & Top1$\uparrow$  & Top5$\uparrow$  & hr$\downarrow$ \\
				\midrule[0.5pt]
				\multirow{2}{*}{-}  
				& Teacher  &100\% & 62.77 & 84.55 & 3.3 & 74.07 & 91.05 &3.7  \\
				& Student& 100\% & 51.08 & 76.05 & 2.6 & 70.81 & 89.61 & 2.9 \\
				\midrule[0.5pt]
				\multirow{3}{*} {\makecell[l]{KD\cite{geoffrey-arxiv2015-kd}\\ \small arXiv'15 }} 
				&Vanilla & 100\%& \underline{54.81} & \underline{78.95} & 2.8& 71.98 & 90.29 &3.1 \\
				& Ours & 10\%& \bf55.26 & \bf79.85 &2.1& \bf73.01& \bf91.00  & 2.4 \\
				& Ours$^\ast$ & 10\% & 54.37 & 78.23 & 0.8 & \underline{72.69} & \underline{90.64} & 1.1 \\
				\midrule[0.5pt]
				\multirow{3}{*}{\makecell[l]{DKD\cite{zhao-cvpr2022-dkd}\\\small CVPR'22}} 
				&Vanilla & 100\%& 56.17 & 80.20 & 2.9 & \underline{73.95} & \underline{92.10} &3.1\\
				& Ours& 10\%& \bf56.74 & \bf81.72 & 2.2 & \bf74.07 & \bf92.13 & 2.3 \\
				& Ours$^\ast$ &10\%& \underline{56.35} & \underline{80.88} & 0.8 & 73.92 & 91.47 &1.0 \\
				\midrule[0.5pt]
				\multirow{3}{*}{\makecell[l]{CTKD\cite{li-aaai2023-ctkd} \\ \small AAAI'23}}
				& Vanilla&100\% & \underline{55.21} & 79.95 & 3.0 & 72.23 & 90.72 & 3.2 \\
				& Ours& 10\%& \bf55.99 & \bf81.02 & 2.3& \bf73.12 & \bf91.13 & 2.3 \\
				& Ours$^\ast$ &10\%& 54.82 & \underline{80.92} & 0.8 & \underline{72.83} & \underline{90.87} &1.1\\
				\midrule[0.5pt]
				\multirow{3}{*}{\makecell[l]{CrossKD\cite{wang-cvpr2024-crosskd}\\\small CVPR'24 }}
				& Vanilla &100\%& {58.07} & 79.26 &2.8& 74.71 & \underline{91.86} & 3.0\\
				& Ours &10\%& \bf{62.18} & \bf81.02 & 2.2 & \bf 75.07 & \bf92.06 & 2.2\\
				& Ours$^\ast$ &10\%& \underline{60.28} & \underline{80.54} & 0.8 & \underline{74.85} & 91.72 & 1.0\\
				\midrule[0.5pt]
				\multirow{3}{*}{\makecell[l]{DualKD\cite{wang-tip2024-dkd} \\ \small TIP'24}}
				& Vanilla &100\% & 56.25 & 79.85 & 3.1 & 74.09 & 91.29 & 3.3\\
				& Ours &10\%& \bf57.02 & \bf81.95 & 2.4 & \bf74.87 & \bf91.82 & 2.5\\
				& Ours$^\ast$ &10\% & \underline{56.37} & \underline{81.83} & 0.9 & \underline{74.28} & \underline{91.39} &1.2  \\
				\toprule[0.75pt]
			\end{tabular}
		}
	}
	\vspace{-3mm}
\end{table}

\begin{table}[!t]
	\centering
	\caption{Performance on CIFAR-100 \cite{wah-2011-cifar} in terms of Top-1/5 Accuracy (\%) / training time (sec). ``$r$'' is sample selection ratio.}
	\label{tbl:cifar}
	\scalebox{0.75}{
		\setlength{\tabcolsep}{0.7mm}{
			\begin{tabular}{llc ccccc}
				\toprule[0.75pt]
				\multirow{3}{*}{Method} &\multirow{3}{*} {Type} & \multirow{3}{*}{$r$} & \multicolumn{2}{c}{WRN \cite{zagoruyko-bmvc2017-residual} } & \multicolumn{3}{c}{ResNet \cite{he-cvpr2016-residual}} \\
				\cmidrule(lr){4-5} \cmidrule(lr){6-8}
				&  & & WRN40-2 & WRN40-2 & ResNet56 & ResNet110 & ResNet32$\times$4 \\
				&  & & WRN16-2 & WRN40-1 & ResNet20 & ResNet32  & ResNet8$\times$4 \\
				\midrule[0.5pt]
				\multirow{2}{*}{-} 
				& Teac. & 100\%& 75.61 (45.2s) & 75.61 (54.2s) & 72.34 (50.3s) & 74.31 (68.2s) & 79.42 (39.8s) \\
				& Stud. & 100\% & 73.26 (38.2s) & 71.09 (46.2s) & 69.06 (43.2s) & 70.87 (60.3s) & 71.50 (33.8s) \\
				\midrule[0.5pt]
				\multirow{3}{*} {\makecell[l]{KD\cite{geoffrey-arxiv2015-kd}\\ \small arXiv'15 }} 
				& Vani. & 100\% & \underline{74.34} (40.2s) & \textbf{73.02} (48.8s) & \bf{70.81} (46.2s) & \underline{73.18} (64.6s) & \underline{72.23} (36.1s) \\
				& Ours & 50\%  & \textbf{74.44} \underline{(33.6s)} & \underline{72.86} \underline{(39.4s)} & \underline{70.62} \underline{(40.6s)} & \textbf{73.36} \underline{(50.8s)} & \textbf{72.85} \underline{(26.3s)} \\
				& Ours$^\ast$ & 50\%  & 74.23 \textbf{(22.8s)} & 72.69 \textbf{(27.4s)} & 70.55\textbf {(26.6s)} & 73.12\textbf{(36.0s)} & 72.05 \textbf{(21.6s)} \\
				\midrule[0.5pt]
				\multirow{3}{*}{\makecell[l]{DKD\cite{zhao-cvpr2022-dkd}\\ \small CVPR'22}} 
				& Vani. &100\% & \textbf{76.17} (40.2s) & \textbf{74.91} (50.4s) &\textbf{71.86} (45.3s) & \underline{73.41} (64.2s) & \underline{75.92} (36.2s) \\
				& Ours & 50\%  & \underline{75.95} \underline{(33.6s)}& \underline{74.88} \underline{(38.2s)} & \underline{71.35} \underline{(35.4s)} & \textbf{73.87} \underline{(48.6s)} & \textbf{76.02} \underline{(27.0s)} \\
				& Ours$^\ast$ & 50\%  & 75.62 \textbf{(22.4s)} & 74.54 \textbf{(27.8s)} & 70.92 \textbf{(25.9s)} & 72.92 \textbf{(35.4s)} & 75.57 \textbf{(21.5s)} \\
				\midrule[0.5pt]
				\multirow{3}{*}{\makecell[l]{CTKD\cite{li-aaai2023-ctkd} \\ \small AAAI'23}}
				& Vani. & 100\% & 74.58 (41.2s) & 72.83 (50.2s) & \underline{71.08} (46.9s) & \underline{72.83} (65.2s) & \underline{72.52} (36.2s) \\
				& Ours & 50\%  & \textbf{74.98} \underline{(35.1s)} & \textbf{72.93} \underline{(33.8s)} & \textbf{70.92} \underline{(41.2s)} & \textbf{72.91} \underline{(53.6s)} & \textbf{72.63} \underline{(26.8s)} \\
				& Ours$^\ast$ & 50\% & \underline{74.83} \textbf{(23.5s)} & \underline{72.85} \textbf{(28.0s)} & 70.64 \textbf{(27.0s)} & 72.62 \textbf{(36.8s)} & 72.20 \textbf{(21.4s)} \\
				\midrule[0.5pt]
				\multirow{3}{*}{\makecell[l]{\small{CrossKD\cite{wang-cvpr2024-crosskd}}\\ \small CVPR'24 }}
				& Vani. & 100\% & 73.41 (40.9s) & \underline{71.33} (49.5s) & \textbf{69.95} (46.2s) & \textbf{71.58} (64.1s) & \underline{71.90} (36.7s) \\
				& Ours & 50\%  & \textbf{74.01} \underline{(36.2s)} & \textbf{71.42} \underline{(41.0s)} & \underline{69.73} \underline{(41.6s)} & \underline{71.42} \underline{(48.5s)} &\textbf{71.98} \underline{(27.4s)}\\
				& Ours$^\ast$ & 50\%  & \underline{73.87} \textbf{(23.6s)} & 71.11 \textbf{(28.0s)} & 69.21 \textbf{(26.8s)} & 71.05 \textbf{(35.3s)} & 71.48 \textbf{(21.3s)} \\
				\midrule[0.5pt]
				\multirow{3}{*}{\makecell[l]{\small{DualKD\cite{wang-tip2024-dkd}} \\ \small TIP'24}}
				& Vani. & 100\% & \textbf{74.24} (41.3s) & \textbf{73.86} (50.3s) & \underline{70.77} (46.8s) & \underline{72.87} (65.3s) & \underline{72.50} (36.9s) \\
				& Ours & 50\% & \underline{74.01} \underline{(34.4s)} & \underline{73.53} \underline{(40.9s)} & \textbf{70.82} \underline{(40.4s)} & \textbf{72.98} \underline{(50.6s)} &\textbf{73.52} \underline{(26.3s)} \\
				& Ours$^\ast$ & 50\% & 73.78 \textbf{(23.4s)} & 73.29 \textbf{(28.3s)} & 70.12 \textbf{(26.8s)} & 72.55 \textbf{(36.2s)} & 72.18 \textbf{(22.3s) }\\
				\toprule[0.75pt]
			\end{tabular}
		}
	}
\end{table}

\begin{table}[!t]
	\centering
	\caption{Ablation studies on components in terms of Top-1/5 Accuracy (\%). ``Diff'' denotes distillation difficulty evaluation module, ``Ada.'' denotes adaptive distillation module. }
	\label{tbl:abla_component}
	\scalebox{0.72}{
		\setlength{\tabcolsep}{0.3mm}{
			\begin{tabular}{l c cc c cc c cc c cc ccc cccccc}
				\toprule[0.75pt]
				\multirow{4}{*}{\small Method} & \multirow{4}{*}{Dif.} & \multirow{4}{*}{Ada.} 
				& &\multicolumn{5}{c}{UCF101} && \multicolumn{5}{c}{Kinetics-400} 
				&& \multicolumn{5}{c}{CIFAR100} & \\
				\cmidrule(lr){5-9} \cmidrule(lr){11-15} \cmidrule(lr){17-21}
				& & & & \multicolumn{2}{c}{SlowFast} && \multicolumn{2}{c}{VideoST} 
				&& \multicolumn{2}{c}{SlowFast} && \multicolumn{2}{c}{VideoST}
				&& \multicolumn{2}{c}{ResNet} && \multicolumn{2}{c}{WRN} &\\
				\cmidrule(lr){4-7} \cmidrule(lr){7-10} \cmidrule(lr){10-13} \cmidrule(lr){13-16}\cmidrule(lr){16-19}\cmidrule(lr){19-22}
				& & & & Top1$\uparrow$ & Top5$\uparrow$ & & Top1$\uparrow$ & Top5$\uparrow$
				&& Top1$\uparrow$ & Top5$\uparrow$ && Top1$\uparrow$ & Top5$\uparrow$ 
				&& Top1$\uparrow$ & Top5$\uparrow$ && Top1$\uparrow$ & Top5$\uparrow$ &\\
				\midrule[0.5pt]
				\multirow{4}{*}{\makecell[l]{KD\cite{geoffrey-arxiv2015-kd}\\ \small  }} 
				& & & & 78.91 & 94.33 && 84.77 & 96.07 && 50.23 & 74.23 && 66.24 & 87.23 && 68.01 & 90.91 && 72.01 & 92.24 & \\
				& \checkmark &  & & 81.74 & 96.10 && \underline{86.50} & \underline{97.72} && \underline{53.01} & \underline{78.27} && \underline{71.24} & \underline{90.23} && \underline{69.25} & \underline{91.80} && 72.95 & 92.79 & \\
				& & \checkmark  & & \underline{88.95} & \underline{98.33} && 86.07 & 96.64 && 51.98 & 76.44 && 70.92 & 90.05 && 68.75 & 91.63 && \underline{73.18} & \underline{92.84} & \\
				& \checkmark & \checkmark & & \bf89.88 & \bf98.80 && \bf87.70 & \bf98.72 && \bf54.37 & \bf78.23 && \bf72.69 & \bf90.64 && \bf70.55 & \bf92.85 && \bf74.23 & \bf92.87 & \\
				
				\midrule[0.5pt]
				
				\multirow{4}{*}{\makecell[l]{\scriptsize{CrossKD}\\ \small \cite{wang-cvpr2024-crosskd} }} 
				& & & & 86.39 & 97.67 && 83.65 & 96.88 && 56.21 & 76.28 && 68.34 & 88.02 && 67.08 & 89.92 && 70.23 & 91.72 & \\
				& \checkmark & & & 88.44 & \underline{99.18} && \underline{85.19} & \underline{97.09} && \underline{57.72} & 77.85 && \underline{72.83} & \underline{91.23} && 67.97 & 90.85 && 71.75 & 92.36 & \\
				&  & \checkmark & & \underline{89.06} & 98.59 && 84.73 & 96.96 && 57.41 & \underline{79.81} && 70.88 & 90.03 && \underline{68.47} & \underline{91.19} && \underline{72.09} & \underline{92.73} & \\
				& \checkmark & \checkmark & & \bf90.04 & \bf99.23 && \bf86.94 & \bf97.16 && \bf60.28 & \bf80.54 && \bf74.85 & \bf92.06 && \bf69.21 & \bf92.93 && \bf73.87 & \bf93.87 & \\
				\toprule[0.75pt]
			\end{tabular}	
		}
	}
\end{table}	

\subsection{Ablation Studies}
To probe into the details of our method, some ablation studies on the individual components, the interruption threshold $\eta$, the distillation strength $\lambda$, and the sample selection ratio $r$ were conducted on UCF101 and CIFAR-100. Here, we use ResNet56/ResNet20 and WRN40-2/WRN16-2 as teacher/student for CIFAR-100. We do ablations using KD \cite{geoffrey-arxiv2015-kd} by selecting samples every five epochs, and hyper-parameters keep still as during training unless specified. Please refer to supplementary for more results. 

\textbf{Individual component}. We examine the sample distillation difficulty (SDD) evaluation module and the sample adaptive distillation (SAD) module, whose results are shown in Table~\ref{tbl:abla_component}, where the baseline without ``$\checkmark$'' is the vanilla KD method with DPP sampling. As shown in the top, using SDD module (row~2) improves the performance by about 2.8\% in terms of Top-1 accuracy with SlowFast model on UCF101, and that improvement with VideoST is larger on Kinetics-400, \ie, 5.0\%; the performance gains are significant (\ie, 10\%) by adding SAD module (row~3) with SlowFast on UCF101; coupling the two module (row~4) achieves the best performance. Similar observations are also found by using CrossKD in the bottom group. This demonstrates considering the sample distillation difficulty and adaptively updating the sample distillation strength are beneficial for transferring knowledge from teacher, regardless of logit-based or feature-based methods.

\textbf{Interruption threshold $\eta$}. We vary $\eta$ from 0 to 1 with seven grids, and show the results in Table~\ref{tbl:abla_interruption_threshold}. As the threshold increases from 0, the performance is progressively enhanced and mostly achieves the best around 0.5, but it tends to degrade after the peak point. This indicates that the threshold should be neither too large nor too small, since it decides whether doing random dropout ($<\eta$) or random shuffle on video frames ($\ge\eta$). 

\textbf{Distillation strength $\lambda$}. We vary $\lambda$ from 0 to 1 with seven grids, and show the results in Table~\ref{tbl:abla_distill_strength_para}. From the table, the performance achieves the best when $\lambda$ takes 0.1, \ie, the previous distillation strength contributes 10\%, while the current one contributes 90\% to the model. This suggests the history distillation strength cannot be neglected and the current one dominates the adaptive distillation process.     

\begin{figure}[!t]
	\centering
	\begin{subfigure}[b]{0.49\linewidth}
		\includegraphics[width=0.95\linewidth]{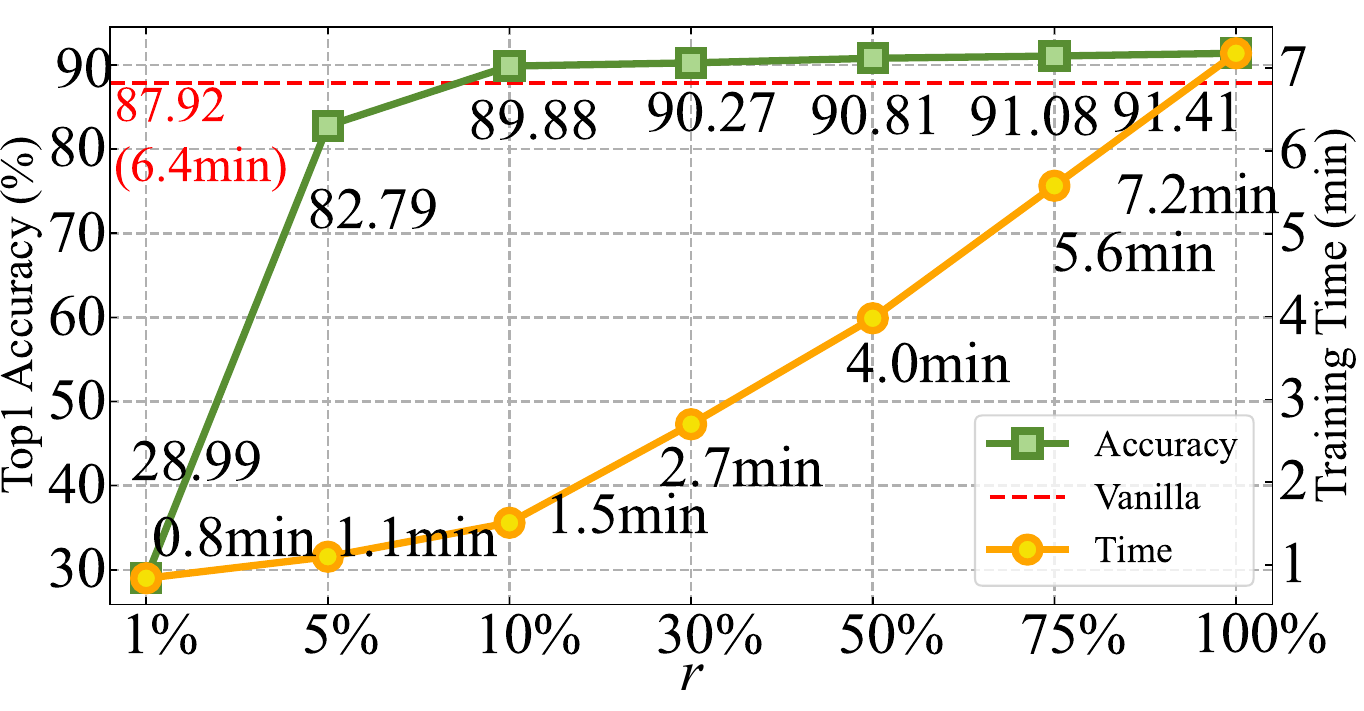}
		\caption{UCF101\cite{soomro-arxiv2012-ucf101}-SlowFast\cite{feichtenhofer-iccv2019-slowfast}}
		\label{fig:vis_top1_slowfast_ucf101}
	\end{subfigure}
	\begin{subfigure}[b]{0.49\linewidth}
		\includegraphics[width=0.95\linewidth]{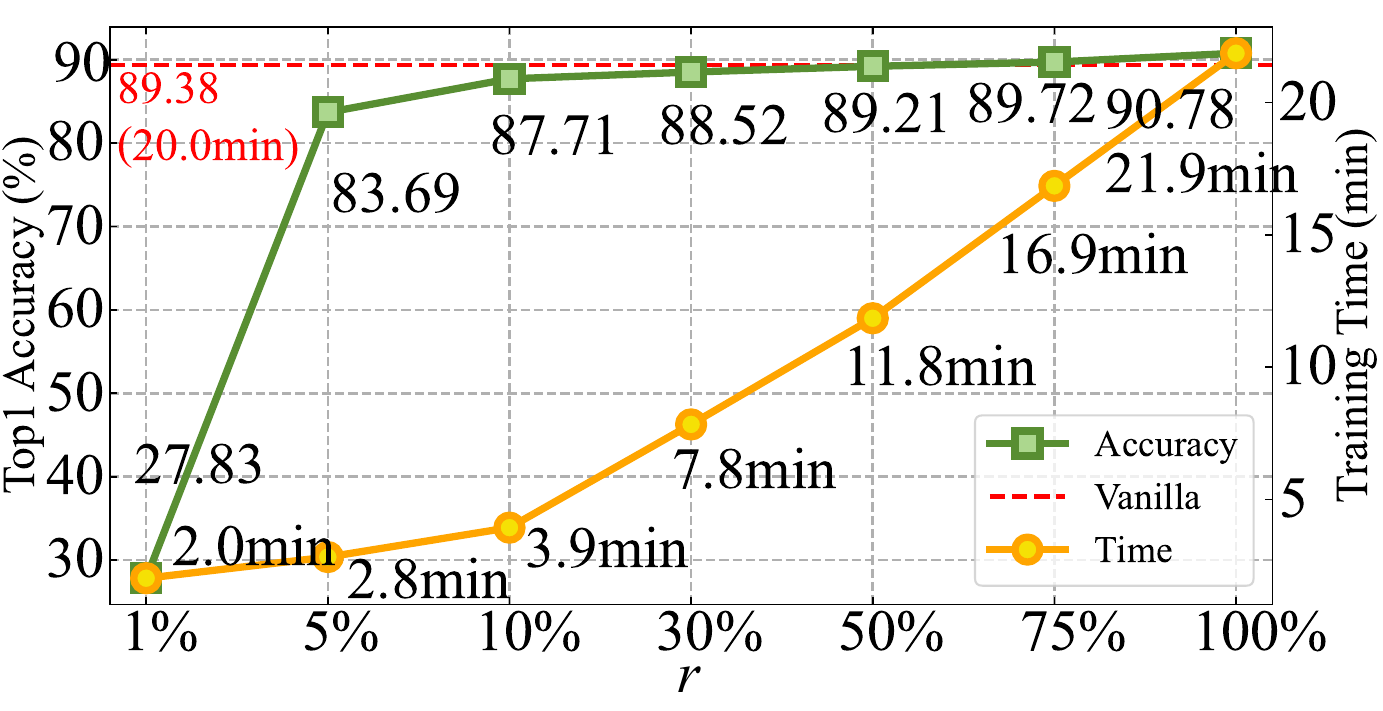}
		\caption{UCF101\cite{soomro-arxiv2012-ucf101}-VideoST\cite{liu-cvpr2022-video}}
		\label{fig:vis_top1_viseost_ucf101}
	\end{subfigure}
	\begin{subfigure}[b]{0.49\linewidth}
		\includegraphics[width=0.95\linewidth]{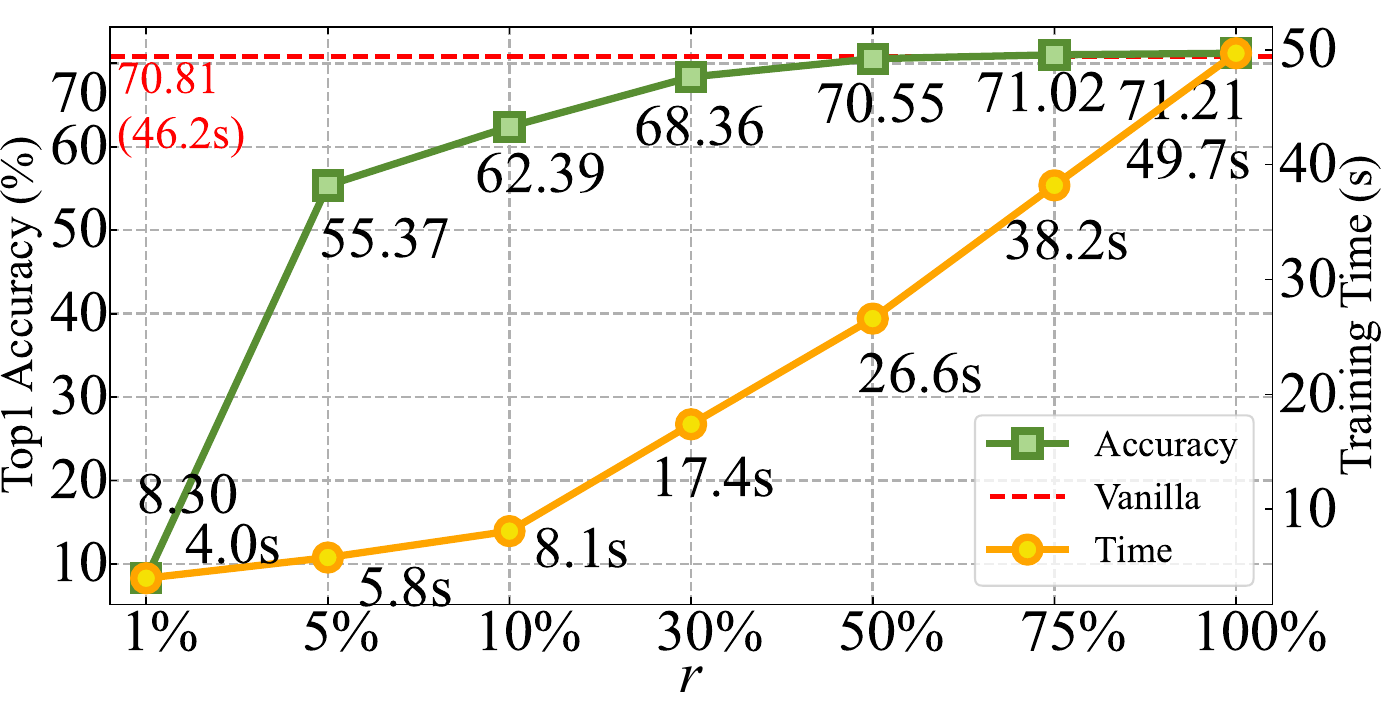}
		\caption{CIFAR-100\cite{wah-2011-cifar}-ResNet\cite{he-cvpr2016-residual}}
		\label{fig:vis_top1_resnet_cifar100}
	\end{subfigure}
	\begin{subfigure}[b]{0.49\linewidth}
		\includegraphics[width=0.95\linewidth]{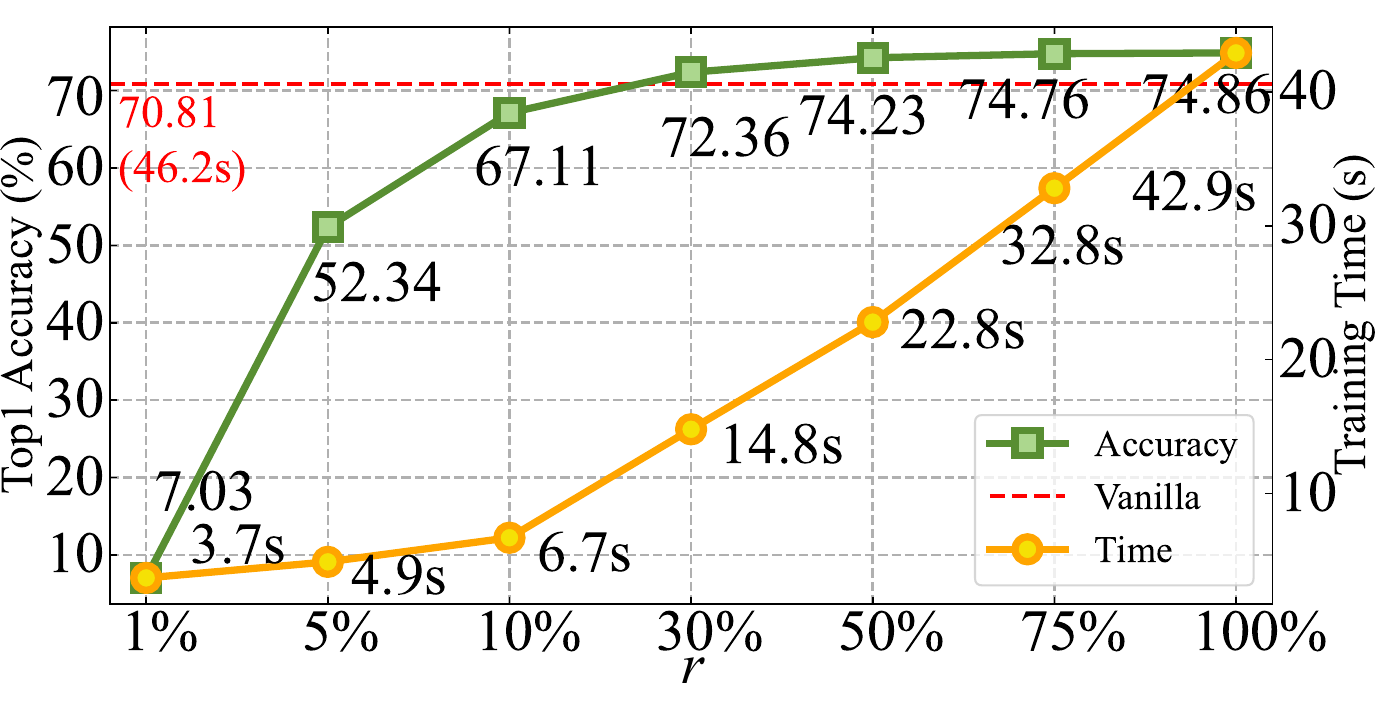}
		\caption{CIFAR-100\cite{wah-2011-cifar}-WRN\cite{zagoruyko-bmvc2017-residual}}
		\label{fig:vis_top1_wrn_cifar100}
	\end{subfigure}
	\caption{Performance vs. sample selection ratio $r$. }
	\label{fig:vis_top1_kd}
\end{figure}

\textbf{Sample selection ratio $r$}. We vary $r$ from 1\% to 100\% and depict the Top-1 accuracy and the training time in Fig.~\ref{fig:vis_top1_kd}. As drawn by the curves, the accuracy rapidly approaches (VideoST) or even surpasses (SlowFast) that of baseline (vanilla KD) by only 10\% of training samples on two video datasets during student learning. The elapsed training time is very low compared to that of baseline using all training samples. Note that CIFAR-100 is an image dataset which is much easier for classification, which makes knowledge distillation more difficult, so it desires almost 50\% (ResNet) or 30\% (WRN) of training samples to achieve that of baseline.

\begin{table}[!t]
	\centering
	\caption{Ablation studies on the interruption threshold $\eta$.}
	\label{tbl:abla_interruption_threshold}
		\setlength{\tabcolsep}{0.4mm}{
			\begin{tabular}{c ccccccc cccccc}
				\toprule[0.75pt]
				\multirow{3}{*}{$\eta$} &&  \multicolumn{5}{c}{UCF101\cite{soomro-arxiv2012-ucf101}} &&\multicolumn{5}{c}{CIFAR100\cite{wah-2011-cifar}} &\\
				\cmidrule(lr{0.5pt}){2-7}  \cmidrule(lr{0.5pt}){8-13}
				&&\multicolumn{2}{c}{Top1$\uparrow$} & &\multicolumn{2}{c}{Top5$\uparrow$} &&\multicolumn{2}{c}{Top1$\uparrow$} & &\multicolumn{2}{c}{Top5$\uparrow$} &  \\ 
				\cmidrule(lr{0.5pt}){2-4}  \cmidrule(lr{0.5pt}){5-7} \cmidrule(lr{0.5pt}){8-10}  \cmidrule(lr{0.5pt}){11-13}
				&& SlowFast	& VideoST && SlowFast	& VideoST && ResNet	&WRN & & ResNet	&WRN&\\ 
				\midrule[0.5pt]
				0.0	&& 88.34	& 86.82		&	&98.43		&  96.59	&		& 69.01			& 73.19	&		&91.52  &93.27&\\
				0.1	&&88.42    &87.01&& 98.67			&97.33	&	& 69.55 	&73.08	&	&91.82 			 			&93.34 & \\
				0.3	&& 89.27 	&\underline{87.52}&	& \underline{98.64}						&  98.05	&		& 70.02			& 73.64		&	&92.37  &93.76&\\
				0.5	&&\bf89.88  &\bf87.70&& \bf 98.80 			&\underline{98.72}&	&\bf70.55		&\bf73.87 & 	 	&\underline{92.85} 		  &\underline{93.79}&\\
				0.7	&& \underline{89.53} 	&87.30&& 98.07				&  	\bf98.84	&	&\underline{70.46} 		&\underline{73.82} 		&	&\bf92.93  &\bf93.86&\\
				0.9	&& 89.14    & 87.03	&		&97.99			&  	97.93	&	&69.32 			& 73.65	&		&92.36  &93.54&\\
				1.0	&& 88.78	&86.87 		&	&96.87		&97.70  	&		& 69.43		&73.76&	& 92.18			  &93.20&\\
				\toprule[0.75pt]
			\end{tabular}
		}
\end{table}			

\begin{table}[!t]
	\centering
	\caption{Ablation on distillation strength hyper-parameter $\lambda$.}
	\label{tbl:abla_distill_strength_para}
		\setlength{\tabcolsep}{0.4mm}{
			\begin{tabular}{c ccccccc cccccc}
				\toprule[0.75pt]
				\multirow{3}{*}{$\lambda$} &&  \multicolumn{5}{c}{UCF101\cite{soomro-arxiv2012-ucf101}} &&\multicolumn{5}{c}{CIFAR100\cite{wah-2011-cifar}} &\\
				\cmidrule(lr{0.5pt}){2-7}  \cmidrule(lr{0.5pt}){8-13}
				&&\multicolumn{2}{c}{Top1$\uparrow$} & &\multicolumn{2}{c}{Top5$\uparrow$} &&\multicolumn{2}{c}{Top1$\uparrow$} & &\multicolumn{2}{c}{Top5$\uparrow$} &  \\ 
				\cmidrule(lr{0.5pt}){2-4}  \cmidrule(lr{0.5pt}){5-7} \cmidrule(lr{0.5pt}){8-10}  \cmidrule(lr{0.5pt}){11-13}
				&& SlowFast	& VideoST && SlowFast	& VideoST && ResNet	&WRN & & ResNet	&WRN&\\ 
				\midrule[0.5pt]
				0.0 & & 89.69       &87.24&&98.64 &	98.03	&		&  69.81			& 	73.20	&		& 92.01			&93.39&\\
				0.1	& &\bf{89.96}  &\bf{87.79}& &\bf 98.80 		&\bf98.72		&	&\bf70.55  			& \bf73.87			&& \bf92.85 &\bf93.97&\\
				0.3	&&\underline{89.91} 	&\underline{87.70}&	&\underline{98.67}		&	\underline{98.35}	&	& \underline{70.23} 			& 73.56	&		& \underline{92.63}		& 93.41 &\\
				0.5	&&89.87	 &	87.65 &&98.52			&  98.49&			& 70.02			&\underline{73.69}	&		&92.45  &93.32&\\
				0.7	&&89.77 &87.28 &	&98.53			& 	98.02&	&69.87 				&73.48 && 	92.32 &\underline{93.67}&\\
				0.9	&& 89.43			& 86.72		&	&98.42		&97.98&	& 68.85 			 			& 72.87		&	&92.18  &93.25&\\
				1.0	&  & 89.38	&86.01		&	&98.36 			&  97.43	&		& 69.16			&73.47 		&	&92.02  &93.59&\\
				\toprule[0.75pt]
			\end{tabular}
		}
\end{table}		

\subsection{Qualitative Results}
\begin{figure}[!t]
	\centering
	\includegraphics[width=1\linewidth]{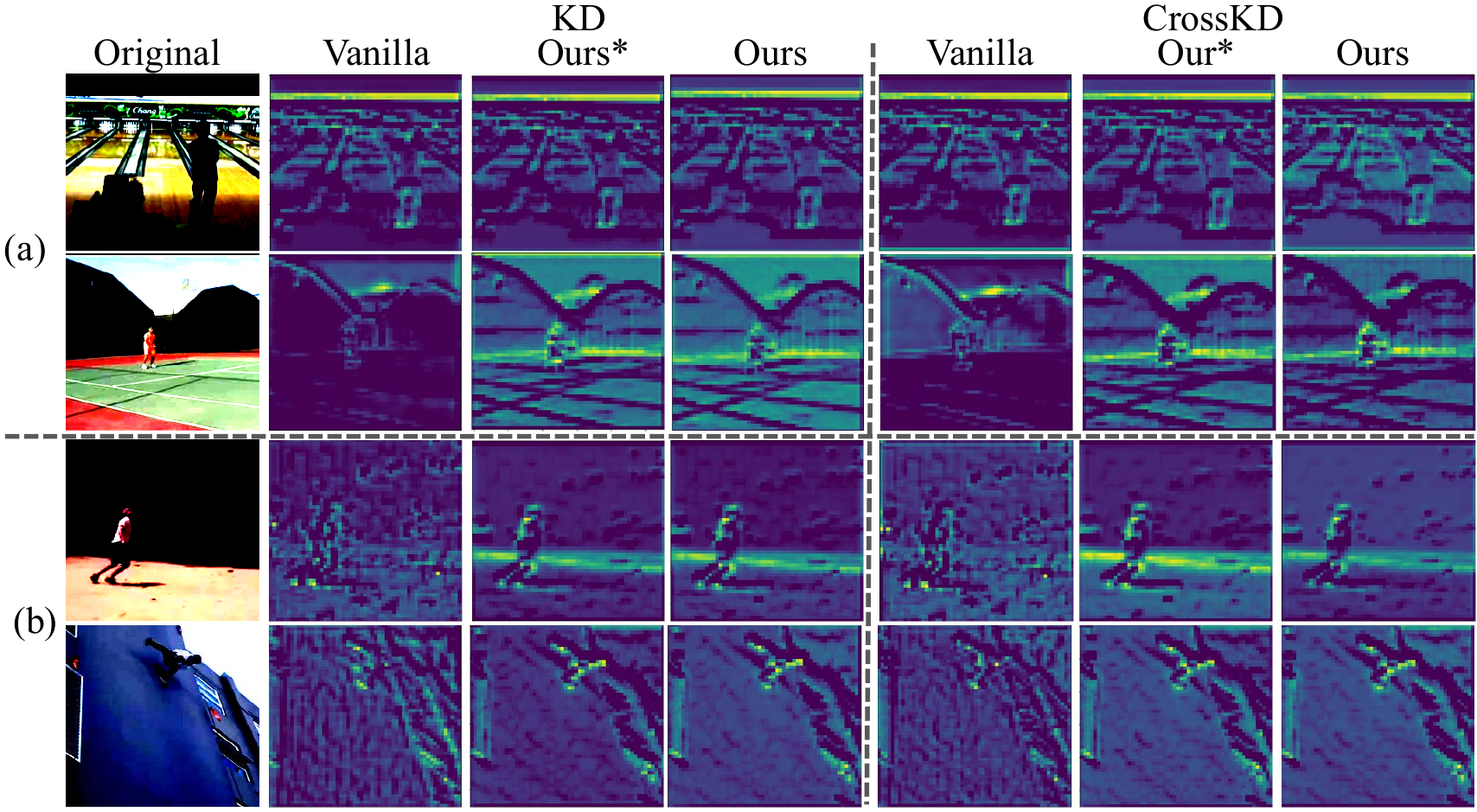}
	\caption{Feature map visualization of our method on SlowFast\cite{feichtenhofer-iccv2019-slowfast}. (a) UCF101 \cite{soomro-arxiv2012-ucf101}; (b) Kinetics-400 \cite{kay-arXiv2017-kinetics}.}
	\label{fig:vis_fea_slowfast}
\end{figure}

To intuitively show the advantage of our method, we randomly choose several videos from UCF101             \cite{soomro-arxiv2012-ucf101} and Kinetics-400 \cite{kay-arXiv2017-kinetics} and visualize the feature maps in Fig.~\ref{fig:vis_fea_slowfast}. These features are from the 2nd block in the slow path of SlowFast \cite{feichtenhofer-iccv2019-slowfast} model by applying our method to vanilla KD \cite{geoffrey-arxiv2015-kd} and vanilla CrossKD \cite{wang-cvpr2024-crosskd}. As shown in Fig.~\ref{fig:vis_fea_slowfast}(a), vanilla KD or CrossKD fails to clearly show the contour in dark area (bottom in row~1 and top in row~2), which may cause a wrong recognition (\eg, ``bowling'' $\rightarrow$ ``standing''). As shown in Fig.~\ref{fig:vis_fea_slowfast}(b), vanilla KD or CrossKD generates the feature maps with many noisy pixels which adds the discriminating difficulty of actions. Our method not only highlights the contours in dark area but also greatly reduces the number of noisy pixels, regardless of selecting samples every epoch (Ours) or every five epochs (Ours$^\ast$). This validates the robustness of our method in adverse condition.

\section{Conclusion}
This work explores the knowledge distillation problem at sample level for action recognition, by developing an efficient adaptive distillation framework. It is inspired by a fact that the transfer difficulty varies across different samples during distillation. In particular, we first evaluate sample distillation difficulty by considering the sample selection probability and the distillation loss of the interrupted samples, after which we calculate the sample distillation strength based on the interruption rate and the distillation difficulty. Meanwhile, we select only a small fraction of samples with both low distillation difficulty and high diversity to train student model. Empirical studies on several benchmarks validate that our method achieves comparable performance or even surpasses prevailing SOTA methods at much lower computational cost.


\ifCLASSOPTIONcaptionsoff
  \newpage
\fi
\end{document}